\documentclass[10pt,twocolumn,letterpaper]{article}
\usepackage{iccv}

\usepackage{times}
\usepackage{epsfig}
\usepackage{graphicx}
\usepackage{amsmath}
\usepackage{amssymb}
\usepackage{enumitem}
\setitemize{noitemsep,topsep=0pt,parsep=0pt,partopsep=0pt}
\usepackage{booktabs}
\usepackage{multirow}
\usepackage{subcaption} %

\usepackage[pagebackref=true,breaklinks=true,letterpaper=true,colorlinks,bookmarks=false]{hyperref}
\usepackage{url}
\makeatletter
\g@addto@macro{\UrlBreaks}{\UrlOrds}
\makeatother
\usepackage{cleveref}

\newcommand{\para}[1]{\noindent \textbf{#1}\xspace}

\newcommand{\Sec}[1]{\S\ref{#1}}

\newcommand{\Fig}[1]{Fig.~\ref{fig:#1}}
\newcommand{\NewPara}[1]{\noindent{\bf #1}}

\newcommand{\eat}[1]{}
\newcommand{\Tab}[1]{Tab.~\ref{tab:#1}\xspace}

\newcommand{\bpp}{bits-per-pixel\xspace}
\newcommand{\sys}{SRVC\xspace}
\newcommand{\ossys}{One-shot Customization\xspace}
\newcommand{\adaptivesys}{\sys\xspace}
\newcommand{\ms}{model stream\xspace}
\newcommand{\cs}{content stream\xspace}
\newcommand{\mb}{model bitrate\xspace}
\newcommand{\cb}{content bitrate\xspace}

\iccvfinalcopy %

\ificcvfinal\fi

\begin{document}
\title{Efficient Video Compression via Content-Adaptive Super-Resolution}

\author{Mehrdad Khani, Vibhaalakshmi Sivaraman, Mohammad Alizadeh\\
MIT CSAIL\\
{\tt\small \{khani,vibhaa,alizadeh\}@csail.mit.edu}
}

\maketitle

\begin{abstract}
    Video compression is a critical component of Internet video delivery. Recent work has shown that deep learning techniques can rival or outperform human-designed algorithms, but these methods are significantly less compute and power-efficient than existing codecs. This paper presents a new approach that augments existing codecs with a small, content-adaptive super-resolution model that significantly boosts video quality. Our method, SRVC, encodes video into two bitstreams: (i) a content stream, produced by compressing downsampled low-resolution video with the existing codec, 
    (ii) a model stream, which encodes periodic updates to a lightweight super-resolution neural network customized for short segments of the video.
    \sys decodes the video by passing the decompressed low-resolution video frames through the (time-varying) super-resolution model to reconstruct high-resolution video frames. Our results show that to achieve the same PSNR, SRVC requires 16\% of the \bpp of H.265 in slow mode, and 2\% of the \bpp of DVC, a recent deep learning-based video compression scheme. SRVC runs at 90 frames per second on a NVIDIA V100 GPU.
\end{abstract}
\section{Introduction}
\begin{figure*}[t]  
  \includegraphics[width=\linewidth]{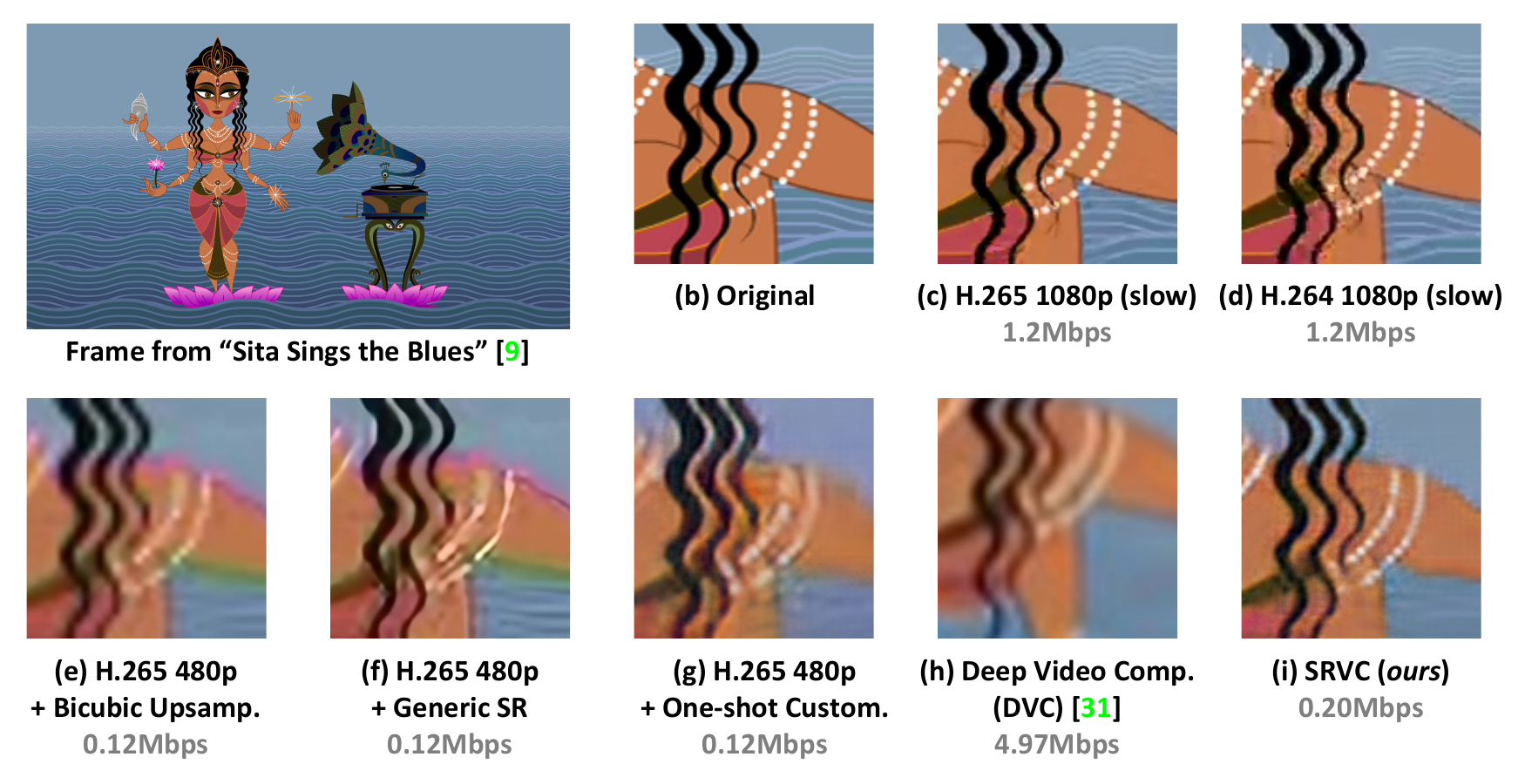}
\caption{Image patches produced by different schemes.}
\label{fig:strip srvc} 
\end{figure*}

Recent years have seen a sharp increase in video traffic. It is predicted that by 2022, video will account for more than 80\% of all Internet traffic~\cite{cisco-video-report,covid-internet-increase}. 
In fact, video content consumption increased so much during the initial months of the pandemic that
content providers like Netflix and Youtube were forced to throttle video-streaming quality to cope
with the surge~\cite{youtube-throttle,amazon-throttle}. 
 Hence efficient video compression 
to reduce bandwidth consumption without compromising
on quality is more critical than ever.

While the demand for video content has increased over the years,
the techniques used to compress and transmit video have largely remained the same. Ideas such as applying Discrete Cosine Transforms (DCTs) to video blocks and computing motion vectors~\cite{H.264, H.263} , which were developed decades ago, are still in use today.
Even the latest H.265 codec improves upon these same ideas by 
incorporating variable block sizes \cite{H.265}. 
Recent efforts \cite{dvc, agustsson2020scale, rippel2019learned} to improve video 
compression have turned to deep learning
to capture the complex relationships between the components of a video compression pipeline.
These approaches have had moderate success at outperforming current codecs,
but they are much less compute- and power-efficient.

We present \sys, a new approach that combines existing compression algorithms with a lightweight, content-adaptive super-resolution (SR) neural network that significantly boosts performance with low computation cost.  \sys compresses the input video into two bitstreams: a \emph{\cs} and a \emph{\ms}, each with a separate bitrate that can be controlled independently of the other stream. The \cs relies on a standard codec such as H.265 to transmit low-resolution frames at a low bitrate. The \ms encodes a {\em time-varying} SR neural network, which the decoder uses to boost the quality of decompressed frames derived from the \cs. \sys uses the \ms to specialize the SR network for short segments of video dynamically (e.g., every few seconds). This makes it possible to use a small SR model, consisting of just a few convolutional and upsampling layers.

Applying SR to improve the quality of low-bitrate compressed video isn't new.  AV1~\cite{chen2018overview}, for instance, has a mode (typically used in low-bitrate settings) that encodes frames at low resolution and applies an upsampler at the decoder. While AV1 relies on standard bicubic~\cite{bicubic} or bilinear~\cite{bilinear} interpolation for upsampling, recent proposals have shown that learned SR models can significantly improve the quality of these techniques~\cite{lin2019improved, feng2018dual}.

However, these approaches rely on {\em generic} SR neural networks~\cite{esrgan,zhang2018unreasonable,isola2017image}) that are designed to generalize across a wide range of input images. These models are large (e.g., 10s of millions of parameters) and can typically reconstruct only a few frames per second even on high-end GPUs~\cite{edsr}. But in many usecases, generalization isn't necessary. In particular, we often have access to the video being compressed ahead of time (e.g, for on-demand video). Our goal is to dramatically reduce the complexity of the SR model in such applications by specializing it (in a sense, overfitting it) to short segments of video.

To make this idea work, we must ensure that the overhead of the \ms is low. Even with our small SR model (with 2.22M parameters), updating the entire model every few seconds would consume a high bitrate, undoing any compression benefit from lowering the resolution of the \cs. \sys tackles this challenge by carefully selecting a small fraction (e.g., 1\%) of parameters to update for each segment of the video, using a ``gradient-guided'' coordinate-descent~\cite{coordesc} strategy that identifies parameters that have the most impact on model quality. Our primary finding is that a SR neural network adapted in this manner over the course of a video can provide such a boost to quality, that including a \ms along with the compressed video is more efficient than allocating the entire bitstream to content.

In summary, we make the following contributions:
\begin{itemize}
    \item We propose a novel dual-stream approach to video streaming that combines a time-varying SR model with compressed low-resolution video produced by a standard codec. We develop a coordinate descent method to update only a fraction of model parameters for each few-second segment of video with low overhead.
    \item We propose a lightweight model with spatially-adaptive kernels, designed specifically for content-specific SR.  Our model runs in real-time, taking only 11 ms (90 fps) to generate a 1080p frame on an NVIDIA V100 GPU. In comparison, DVC~\cite{dvc} takes 100s of milliseconds at the same resolution.
    \item We show that to achieve similar PSNR, \sys requires 16\% of the bits-per-pixel consumed by H.265 in {\em slow mode} 
    \footnote{To the authors' knowledge, this is the first learning-based scheme that compares to H.265 on its slow mode}, 
    and 2\%  of DVC's \bpp. \sys's quality improvement extends across all frames in the video.
\end{itemize}

\Cref{fig:strip srvc} shows visual   examples   comparing the  SRVC with these  baseline approaches at competitive or higher bitrates. Our datasets and code are available at {\url{https://github.com/AdaptiveVC/SRVC.git}}

\begin{figure*}
    \centering
    \includegraphics[width=0.9\linewidth]{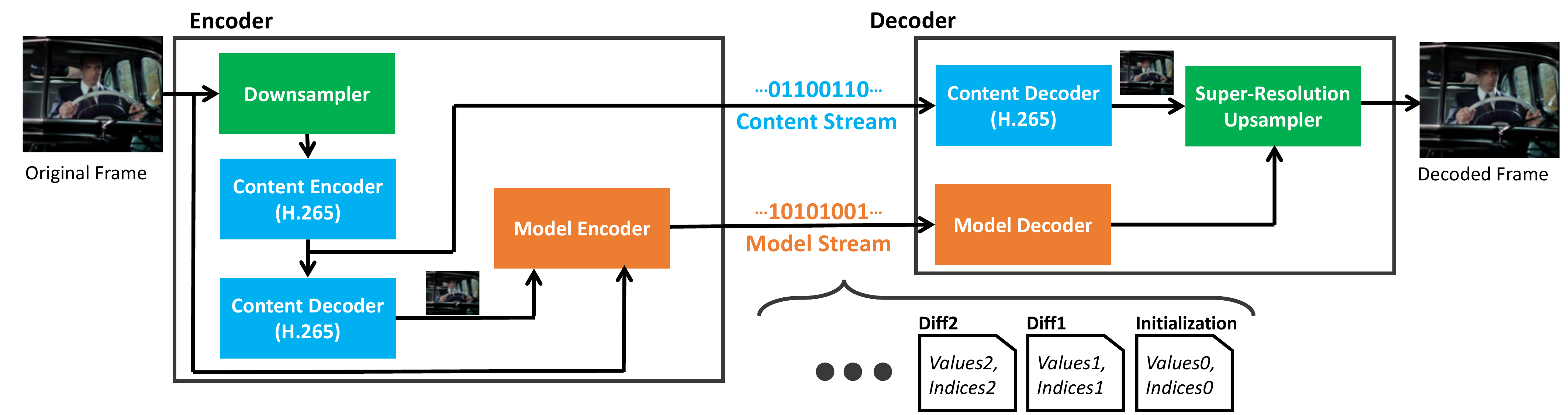}
    \caption{\sys video compression pipeline overview.}
    \label{fig:sys_overview}
\end{figure*}

\section{Related Work}
\para{Standard codecs.} Prior work has widely studied video encoder/decoders (codecs) such as H.264/H.265~\cite{schwarz2007overview, sullivan2012overview}, VP8/VP9~\cite{bankoski2011technical, mukherjee2015technical}, and AV1~\cite{chen2018overview}. These codecs rely on hand-designed algorithms that exploit the temporal and spatial redundancies in video pixels, but cannot adapt these algorithms to specific videos. Existing codecs are particularly effective when used in \emph{slow} mode for offline compression.
Nevertheless, \sys's combination of a low-resolution H.265 stream with a content-adaptive SR model outperforms
H.265 at high resolution, even in its slow mode.
Some codecs like AV1 provide the option to encode at low resolution and upsample using bicubic interpolation~\cite{bicubic}. %
But, as we show in \Sec{sec:eval}, \sys's learned model 
provides a much larger improvement in video quality compared to bicubic interpolation.

\para{Super resolution.} Recent work on single-image SR~\cite{zhang2018unreasonable,isola2017image} and video SR~\cite{lin2019improved,feng2018dual,haris2019recurrent,li2019fast} has produced a variety of CNN-based methods that outperform classic interpolation methods such as bilinear~\cite{bilinear} and bicubic~\cite{bicubic}. Accelerating these SR models has been of interest particularly due to their high computational complexity at higher resolutions~\cite{cvprw_2017_zhang_fast}. Our design adopts the idea of subpixel convolution~\cite{shi2016real}, keeping the spatial dimension of all layers identical to the low-resolution input until the final layer. Fusing the information from several video frames has been shown to further improve single-image SR models~\cite{wang2019edvr}. However, to isolate the effects of using a content-adaptive SR model, we focus on single-image SR in this work. 

\para{Learned video compression.} %
End-to-end video compression techniques~\cite{rippel2019learned,lu2019dvc,agustsson2020scale} follow a compression pipeline similar to standard codecs but replace some of the core components with DNN-based alternatives, e.g., flow estimators~\cite{dosovitskiy2015flownet} for motion compensation and auto-encoders~\cite{habibian2019video} for residue compression. However, running these models in real time is challenging.
For example, even though the model in~\cite{rippel2019learned} is explicitly designed for low-latency video compression, it decodes only 10 frames-per-second (fps) 640$\times$480 resolution on an NVIDIA Tesla V100~\cite{rippel2019learned}. In contrast, H.264 and H.265 process a few 100 frames a second at the same resolution. Moreover, existing learned video compression schemes are designed to generalize and are not targeted to specific videos. %
In this work, we show that augmenting existing codecs with content-adaptive SR achieves better quality and compression than end-to-end learned compression schemes. %

\para{Lightweight models.} 
Lightweight models intended for phones and compute-constrained devices have been designed manually~\cite{MobileNet-v2} and using neural architecture search techniques~\cite{zoph2016neural,wu2019fbnet}.  Model quantization and weight pruning~\cite{he2018amc,lin2016fixed,blalock2020state,CompressionSurvey} have helped reduce the computation footprint of models with a small loss in accuracy. Despite the promise of these optimizations, the accuracy of these lightweight models falls short of state-of-the-art solutions. \sys is complementary to such optimization techniques and would benefit from them.

\section{Methods} \label{sec:method}

\Cref{fig:sys_overview} shows an overview of \sys. \sys compresses video into two bitsreams:  
\begin{enumerate}
    \item {\bf Content stream:} The encoder downsamples the input video frames by a factor of $k$ in each dimension (e.g., $k$=4) to generate low-resolution (LR) frames
    using area-based downsampling. It then encodes the LR frames using an off-the-shelf video codec to generate the content bitstream (our implementation uses H.265~\cite{H.265}). The decoder decompresses the content stream using the same codec to reconstruct the LR frames. Since video codecs are not lossless, the LR frames at the decoder will not exactly match the LR frames at the encoder.  
    
    \item {\bf Model stream:} A second bitstream encodes the SR model that the decoder uses to upsample the each decoded LR frame. We partition the input video into $N$ fixed-length segments, each $\tau$ seconds long (e.g., $\tau=5$). For each segment $t\in\{0,...,N-1\}$, we adapt the SR model to the frames in that segment during encoding. Specifically, the encoder trains the SR model to map the low-resolution decompressed frames within a segment to high-resolution frames. Let $\mathbf{\Theta}_t$ denote the SR model parameters obtained for segment $t$. The model adaptation is sequential: the training procedure for segment $t$ initializes the model parameters to $\mathbf{\Theta}_{t-1}$. The \ms encodes the sequence $\mathbf{\Theta}_{t}$ for $t\in\{0,...,N-1\}$.  It starts with the full model $\mathbf{\Theta}_0$, and then encodes the {\em changes} in the parameters for each subsequent model update, i.e., $\mathbf{\Delta}_t = \mathbf{\Theta}_t - \mathbf{\Theta}_{t-1}$. The decoder updates the parameters every $\tau$ seconds, using the last model parameters $\mathbf{\Theta}_{t-1}$ to find  $\mathbf{\Theta}_{t} = \mathbf{\Theta}_{t-1} + \mathbf{\Delta}_t$. 
    
\end{enumerate}

The \ms adds overhead to the compressed bitstream. To reduce this overhead, we develop a small model that is well-suited to \emph{content-specific} SR (\S\ref{sec:arch}), and design an algorithm that significantly reduces the overhead of model adaptation by training only a small fraction of the model parameters that have the highest impact on the SR quality in each segment (\S\ref{sec:model_stream}).

\subsection{Lightweight SR Model Architecture}\label{sec:arch}

Existing SR models typically use large and deep neural networks (e.g., typical EDSR has 43M parameters across more than 64 layers~\cite{edsr}), making them difficult to use within a real-time video decoder. Moreover, adapting a large DNN model to specific video content and transmiting it to the decoder would incur high overhead.

\begin{figure}
    \centering
    \includegraphics[width=\linewidth]{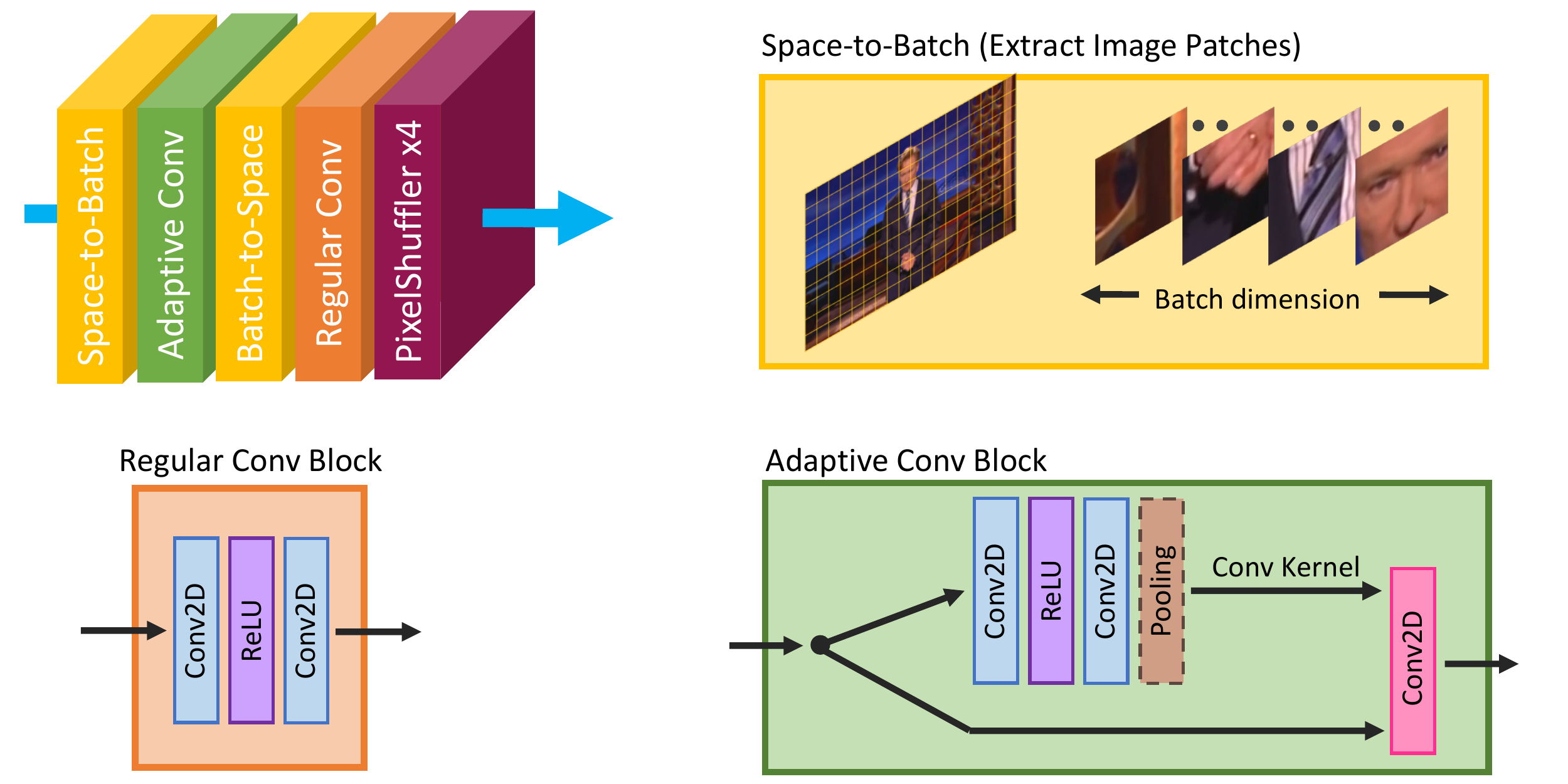}
    \caption{\sys lightweight SR model architecture.}
    \label{fig:arch}
\end{figure}

We propose a new lightweight architecture that keeps the model small and shallow, and yet, is very effective with content-based adaptation (\S\ref{sec:eval results}). Our model is inspired by classical algorithms like bicubic upsampling~\cite{keys1981cubic}, which typically use only one convolutional layer
and a fixed kernel for upsampling the entire image. It uses this basic architecture but replaces the fixed kernel with spatially-adaptive kernels that are customized for different regions of the input frame. Our model partitions each frame into patches, and uses a shallow CNN operating on the patches to generate  different (spatially-adaptive) kernels for each patch. \Fig{arch} shows the architecture.

More formally, the model first partitions an input frame into equal-sized patches of $P \times P$ pixels (e.g. $P=5$ pixels) using a common space-to-batch operation. For each patch, a patch-specific block (Adaptive Conv Block in \Fig{arch}) computes a 3$\times$3 convolution kernel with 3 input and $F$ output channels (27$F$ parameters) using a two-layer CNN, and applies this kernel (the pink box) to the patch.  Therefore, the forward pass of 
the adaptive conv block with input patch $\mathbf{x}\in \mathbb{R}^{P \times P \times 3}$ and output features $\mathbf{y}\in \mathbb{R}^{P \times P \times F}$ is summarized as follows: 
\begin{align*}
    \mathbf{w} &= f(\mathbf{x}),\\
    \mathbf{y} &= \sigma(\mathbf{w} \ast \mathbf{x}).
\end{align*}
We use a two-layer CNN to model $f(\cdot)$ in our architecture.  We finally reassemble the feature patches (batch-to-space) and compute the output using another two-layer CNN followed by a pixel shuffler (depth-to-space)~\cite{shi2016real} that brings the content to the higher resolution. All convolutions have a kernel height and width of 3, except for the first layer of the regular block that uses kernel size of 5.

\subsection{Model Adaptation and Encoding}\label{sec:model_stream}
\para{Training algorithm.} We use the L2-loss between the SR model's output and the corresponding high-resolution frame (input to the encoder), over all the frames in each segment to train the model for that segment.
Formally, we define the loss as
\begin{align*}
    L(\Theta_t) &= \frac{1}{n|F_t|}\sum_{i = 1}^n\sum_{j = 1}^{|F_t|}||Y_{ij} - X_{ij}||^2
\end{align*}
where $|F_t|$ is the number of frames in the $t^{th}$ segment, each with $n$ pixels, and $Y_{ij}$
and $X_{ij}$ denote the value of the $i^{th}$ pixel in the $j^{th}$ frame of the decoded
high-resolution output frame and the original high-resolution input frame respectively.
During the training, we randomly crop the samples at half of their size in each dimension.  We use Adam optimizer~\cite{kingma2014adam} with learning rate of 0.0001, and first and second momentum decay rates of 0.9 and 0.999. 

To reduce the \ms bitrate, we update only a fraction of the model parameters across video segments. Our approach is to update only those parameters that have the maximum impact on the model's accuracy. Specifically, we update the model parameters with {\em the largest gradient magnitude} for each new segment as follows. First, we save a copy of the model at the beginning of a new segment and perform one iteration of training over all the frames in the new segment. We then choose the fraction $\eta$ of the parameters with the largest magnitude of change in this iteration, and reset the model parameters to the starting saved copy. Having selected the set of parameters,  we apply the Adam updates for only these paramaters and discard the updates for the rest of the model (keeping those parameters fixed).

\para{Encoding the \ms.} To further compress the \ms, we only transmit changes to the model parameters at each update. %
We encode the model updates into a bitstream by recording the indices and associated change in values of the model parameters (\Fig{sys_overview}).
\sys's model encoding is lossless: the encoder and decoder both update the same subset of parameters during each update. To update  a fraction $\eta$ of the parameters for a model with $M$ float16 parameters, we need an average bitrate of at most $(16+\log(M))\times \eta M / \tau$ to express the deltas and the indices every $\tau$ seconds. %
For example, with model size $M=2.22$ million parameters ($F$=32, see~\Cref{tab:model speed}), $\tau=10$ seconds, and $\eta = 0.01$, we only require 82~Kbits/sec to encode the \ms
required to generate 1080p video.
To put this number into perspective, Netflix recommends a bandwidth of 5~Mbits/sec at 1080p resolution~\cite{netflix_rec}. The \ms can be compressed further using lossy compression techniques or by dynamically varying $\eta$ or the model update frequency based on scene changes. 

Training the SR model for 1080p resolution and encoding the updates into the \ms takes about 12 minutes for each minute worth of video with our unoptimized implementation. However, given the small compute overhead of our lightweight model, we shared a V100 GPU between five simultaneous model training (encoding) processes without any significance slow down in any of the processes compared to running in single. Hence, the overall throughput of the encoding on V100 GPU is more near 2.5 minutes of training per each one minutes of content. We consider this duration feasible for offline compression scenarios where videos are available to content providers well ahead of viewing time.
However, we believe that there is significant room to accelerate the encoding process with standard techniques (e.g., training on sampled frames rather than all frames) and further engineering. 
We leave an exploration of these opportunities to future work. 

\section{Experiments}
\label{sec:eval} 
\subsection{Setup}
\label{sec:eval setup}

 \begin{figure*}[ht]%
    \centering
    \includegraphics[width=\textwidth]{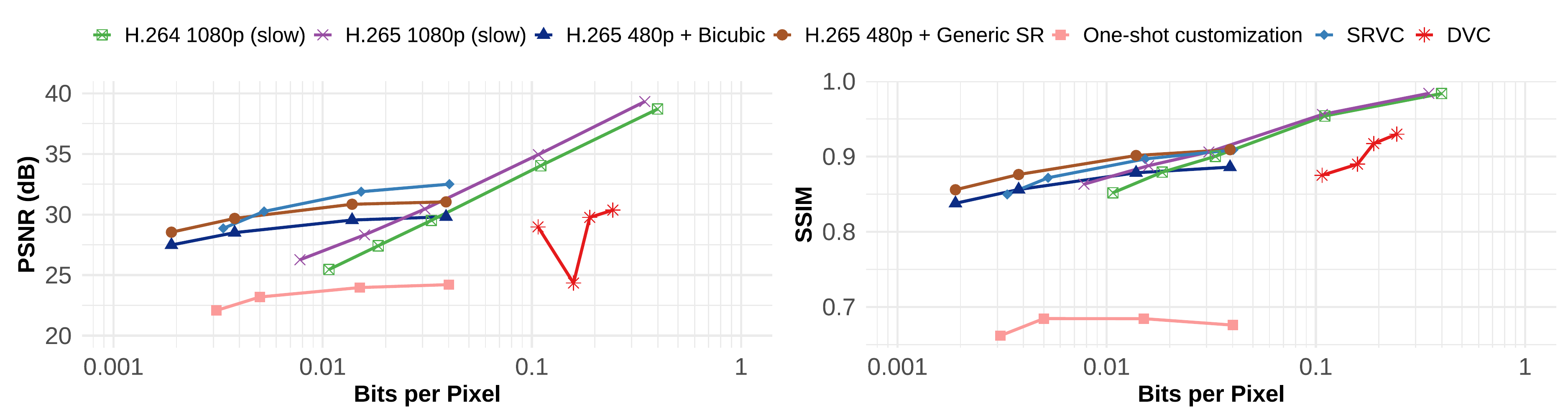}
     \caption{\small Tradeoff between video quality and \bpp for different approaches on three long videos from the Xiph dataset. \sys with content-adaptive streaming reduces the bitrate consumption to 16\% of current codecs 
     and $\sim$2\% of end-to-end compression schemes like DVC. Though comparable in video quality to \sys, the generic SR approach does not run in real-time.}
    \label{fig:quality bpp xiph}
\end{figure*}

\para{Dataset.} Video datasets like JCT-VC~\cite{jctvc}, UVG~\cite{uvg} and MCL-JCV~\cite{wang2016mcl}, consisting of only a few hundred
frames ($\sim$10 sec) per video, are too short to evaluate \sys's content-adaptive SR technique. Hence, we train and test the efficacy of \sys on a custom dataset\footnote{For viewing the videos and more visual examples, see https://github.com/AdaptiveVC/SRVC.git} consisting of 
28 downloadable videos from Vimeo (short films) and 4 full-sequence videos from the Xiph Dataset~\cite{xiph}.
We trim all videos to 10 minutes and resize them %
to 1080p resolution in RAW format from their original 4K resolution and MPEG-4 format using area-based interpolation~\cite{area-interpolation}.
We use the resulting 1080p frames as our high-resolution source frames in our pipeline. We %
re-encode
each video's raw frames at different qualities or Constant Rate Factors (CRFs) on H.264/H.265 
to control the bitrate. 
We also use area-interpolation to 
downsample the video to 480p and encode the low-resolution video using H.265
at different CRFs to %
achieve different degrees of compression.
The SR model in \sys is then trained to learn the mapping from 
each low-resolution video at a particular compression level
to the original 1080p video at its best quality.

\para{Baselines.} 
We compare the following approaches. The first four only use a \cs while the next three
use both a \cs and a \ms. The last approach is an end-to-end neural compression scheme.
\begin{itemize}
\item \textbf{1080p H.264:} We use ffmpeg and the libx264 codec to re-encode each of the 1080p videos at  different compression levels using the {\emph{slow}} preset.
\item \textbf{1080p H.265:} We use ffmpeg and the libx265 codec to re-encode each of the 1080p videos at different compression levels using the {\emph{slow}} preset.
\item \textbf{480p H.265 + Bicubic upsampling:} We use ffmpeg and the libx265 codec to downsample the 1080p original video to its 480p low resolution counterpart at different CRFs using area-interpolation and the {\emph{slow}} preset. This approach 
only uses a \cs: the downsampled 480p frames encoded using H.265. Thus, its bitrate is calculated
based on the downsampled video. We use bicubic 
interpolation to upsample the 480p videos back to 1080p. 
This shows the reduction in bitrates provided by merely encoding at lower-resolutions.
\item \textbf{480p H.265 + Generic SR:} Instead of Bicubic upsampling, we use a more sophisticated DNN-based super-resolution model (EDSR~\cite{edsr}  with 16 residual blocks) to upsample the 480p frames to 1080p. The upsampling takes about 50ms for each frame (about five times more than SRVC). We use a pre-trained checkpoint that has 
been trained on a generic corpus of images~\cite{edsrdataset}. Since we anticipate
all devices to be able to pre-fetch such a model, this approach only has a \cs at 480p encoded
using H.265
and no \ms. %
Thus, its \bpp value is identical to the Bicubic case. 

\item \textbf{480p H.265 + \ossys:} A version of \sys that only uses our
    lightweight SR model (\S\ref{sec:arch})
    without the model adaptation procedure.
    For this, we train our SR model exactly once (one-shot)
        using the entire 1080p video and encode it in the \ms right at the beginning before any low-resolution content.
        The \cs for this approach
        comprises of the 480p H.265 video while the \ms consists of a single initial model 
        customized to the entire video duration. The overhead of the model stream model  
        is amortized over the 
        entire video and added to the content bitrate when computing the total \bpp value. 
\item \textbf{480p H.265 + \adaptivesys:} 
To show the benefits of our content-specific model adaptation procedure, we evaluate \sys 
    which uses the same initial SR model as \ossys but is periodically adapted to the most recent 5 second segment of the video.
        To train this model, we use random crops (half the frame size in each dimension) from each reference frame within a video segment. 
        The \cs for \adaptivesys relies on standard H.265 and thus, its bitrate depends on the compression settings for H.265.
        The \ms, on the other hand, is updated every $5$ seconds and is computed using our gradient-guided strategy, which only encodes the change to those parameters that have the largest gradients in each video segment (\S\ref{sec:model_stream}). To compute the total \bpp, we add the model stream's bitrate (computed as described in \S\ref{sec:model_stream}) to the content stream's bitrate.
        We also add the overhead of sending the initial model in full to the model stream's bitrate.
        
\item \textbf{DVC:} An official checkpoint \cite{dvccheckpoint} of Deep Video Compression \cite{dvc}, an end-to-end neural network based compression algorithm.
To evaluate DVC, we compute the PSNR and SSIM metrics, and use Lu~\etal's\cite{dvc} estimator to measure their required bits-per-pixel for every frame at four different bitrate-distortion trade-off operating points ($\lambda\in\{256,512,1024,2048\}$). 
\end{itemize}

\para{Model and training procedure.}
Our model uses 32 output feature channels in the adaptive convolution block, which results in 2.22 million
parameters. However, note that only 1\% of them are updated by the model stream and that too, only every 5 seconds. We vary the the number of output feature channels, the %
the fraction of model parameters updated, and the update interval 
to understand its impact on \sys's performance.

\para{Metrics and color space.}
We compute the average
Peak Signal-To-Noise Ratio (PSNR) and Structural Similarity Index Measure (SSIM) across all frames at the output of the decoder (after upsampling). We report PSNR based on the mean square error across all
pixels in the video (over all frames) where the pixel-wise error itself 
is computed on the RGB space.
SSIM is computed as the average SSIM between the decoded frames and their corresponding
high-resolution original counterparts.
However, since variations in frame quality over the course of a video can have significant impact on users' experience,
we also show a CDF of both PSNR and SSIM across all frames in the video.

We compute the content bitrate for the all approaches relying on H.264/5 at 
both 1080p and 480p 
using ffmpeg. For approaches that stream a model in addition to video frames,
we compute the model stream bitrate based on the total number of model parameters, the fraction of them that are 
streamed in each update interval, and the
frequency of updates (\S\ref{sec:model_stream}).
The content and model stream bitrates are combined to compute a single \bpp
metric. Note that the \bpp range in our evaluations is an order-of-magnitude lower than results reported in prior work~\cite{dvc, agustsson2020scale}
because we compare to the {\em slow mode} in H.264/5 which is much more efficient than the ``fast'' and ``medium'' modes considered in these papers. We plot PSNR and SSIM metrics at different \bpp to compare different schemes.  
Since \sys upsamples or runs inference on decoded frames as they are rendered to end-users, 
its SR model
needs to run in real-time. To evaluate its feasibility, we also compute \sys's speed in frames per second and compare it to other learning-based approaches.

\subsection{Results}
\label{sec:eval results}

 \begin{figure*}[t]%
    \centering
    \includegraphics[width=\textwidth]{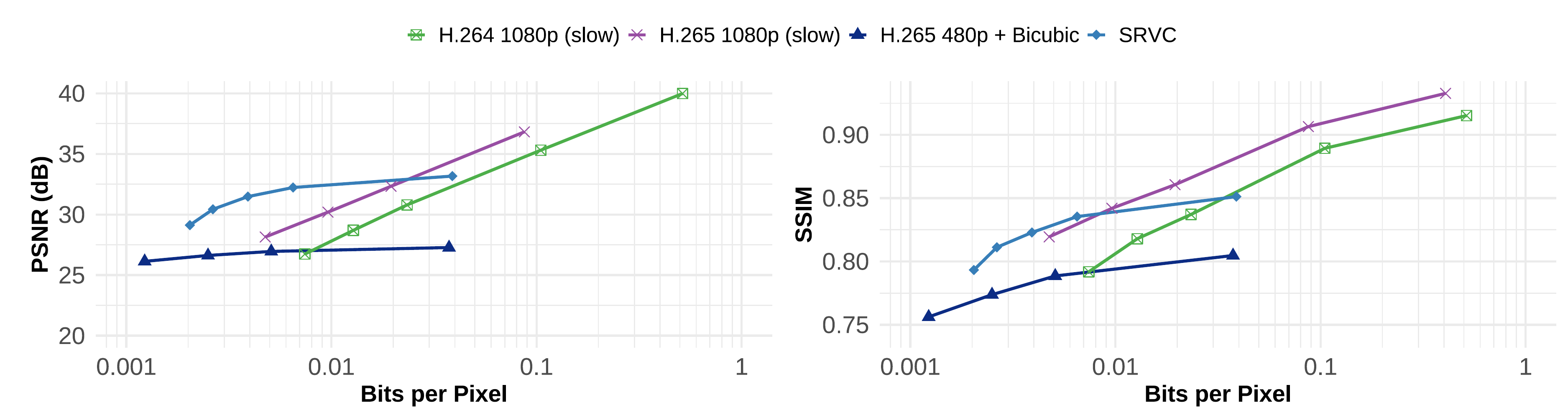}
     \caption{\small  Tradeoff between video quality and \bpp for different approaches on 28 videos from Vimeo. To achieve 30dB PSNR, \sys requires 10\% and 25\% of the \bpp required by H.264 and H.265.}
    \label{fig:quality bpp vimeo}
\end{figure*}

\NewPara{Compression performance.}
\Fig{qualitative strip} shows a visual comparison of the different schemes
for a comparable \bpp value. However, H.264/5 \bpp values can grow significantly 
higher than \sys just due to the range that they operate in. To compare
the compression provided by different approaches across a wider range of \bpp values, we analyzing the PSNR and SSIM achieved by different methods on three long Xiph~\cite{xiph} videos in \Fig{quality bpp xiph}. 
Note that the \bpp metric captures both the contribution of the content and the model for those approaches
that use a \ms for SR.

As in \Fig{quality bpp xiph}, \adaptivesys achieves PSNR comparable to today's H.265 standards with only 16\% of the \bpp.
For instance, to achieve a PSNR of 30 dB, \adaptivesys requires only 0.005 \bpp while H.265 and H.264 codecs, even in their slowest settings,
require more than 0.03 \bpp.
However, \ossys's performs poorer than a simple bicubic interpolation, in alignment with the overfitting goal that we had in designing our custom model. This is because \sys's custom SR model is not large enough to generalize to the entire video, but has enough parameters to learn a small segment.  It is worth noting that to achieve the same PSNR, \adaptivesys requires only 2\% of the \bpp required by DVC~\cite{dvc}, the 
end-to-end neural compression scheme. \adaptivesys's SSIM is comparable
but 0.01-0.02 better than current codecs for the same level of \bpp, particularly at higher bitrates. 

\Fig{quality bpp xiph} suggests that a 480p stream augmented with a generic
SR model performs just as well as \sys in terms of its PSNR and SSIM for a given
\bpp level. However, typical SR models are too slow to perform inference on a single frame (about 5$\times$ slower in this case), making them unfit for real-time video delivery to viewing clients. To evaluate the performance of viable schemes with reasonable video quality on real-world video, we evaluate the \bpp vs. video quality tradeoff on 28 videos publicly available on Vimeo. As \Fig{quality bpp vimeo} suggests, \adaptivesys
outperforms all other approaches on the PSNR achieved for a given \bpp value. In particular, to achieve 30dB PSNR, \adaptivesys requires 25\% and 10\% of the \bpp required by H.265 and H.264 respectively.

\begin{figure}
    \includegraphics[width=\columnwidth]{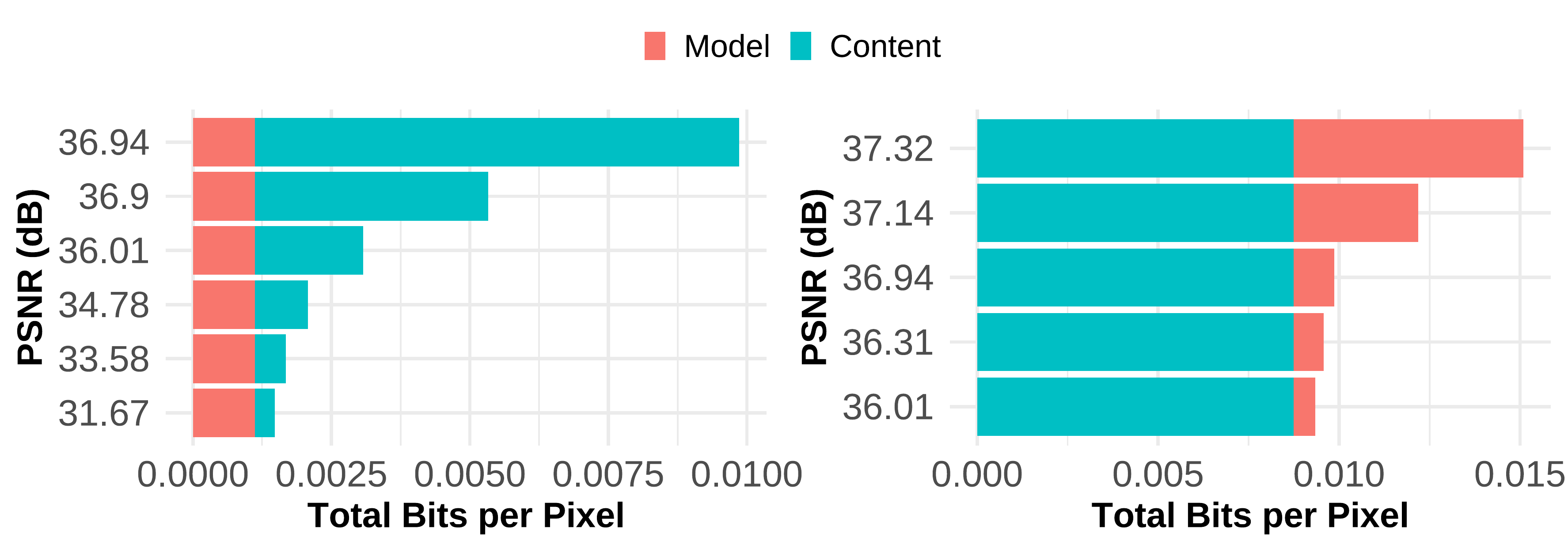}
     \caption{\small Impact of varying \bpp for the \cs for a fixed \mb and vice-versa. 
     Increasing the \bpp for the low-resolution H.265 \cs %
     improves PSNR, especially at low bitrates.
     At a higher \cb, increasing the \bpp dedicated to the \ms by transmitting more model parameters 
     further improves PSNR.}
    \label{fig:model frame ablation}
\end{figure}

A key takeaway from Figures.~\ref{fig:quality bpp xiph} and ~\ref{fig:quality bpp vimeo} is that for a given bitrate budget or level of compression, \sys achieves better
 quality than standard codecs. This suggests that beyond a baseline bitrate for the content, it is better 
to allocate bits to streaming a SR model than to dedicate more bits to the content. We describe this trade-off
between model and content bitrates in more detail in  \Fig{model frame ablation}.

\begin{figure}
    \includegraphics[width=\columnwidth]{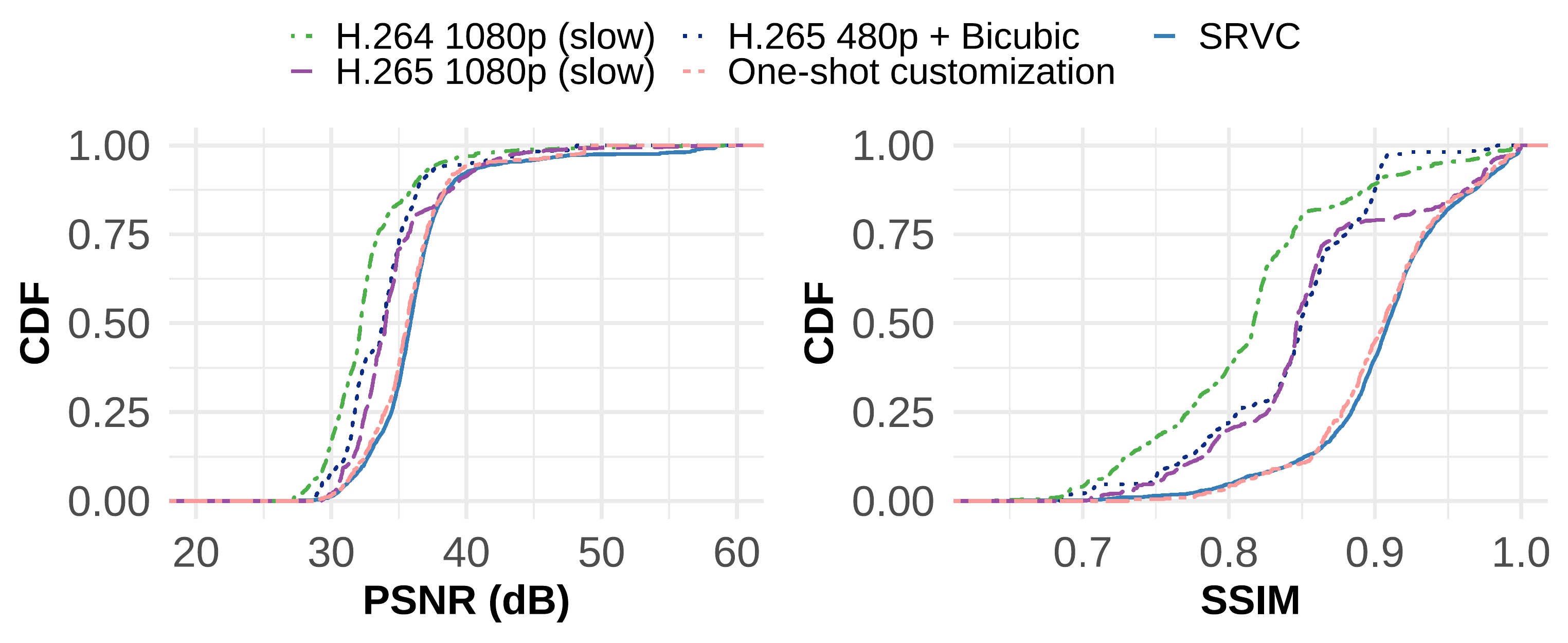}
     \caption{\small CDF of PSNR and SSIM improvements with \sys across all frames in the video at a \bpp of 0.002.
     The quality enhancement from \sys is not limited to only those frames that occur right after a model update, but is
     rather spread out across the entire video.}
    \label{fig:tail quality}
\end{figure}

\NewPara{Robustness of quality improvements.}
To see if \sys's improvements come from just producing a few high-quality frames right after  the model is updated, 
we plot a CDF of the PSNR and SSIM values across all frames of the Meridian video in \Fig{tail quality}.
We compare schemes at a \bpp value of $\sim$0.002. Since DVC~\cite{dvc} has a much higher \bpp and EDSR~\cite{edsr}
performs poorly relative to other approaches, we exclude both approaches.
Firstly, we notice that both \ossys and \adaptivesys perform better than other schemes. Further, this improvement occurs over all of the frames in that
no frame is worse off with \sys than it is with the defacto codec. In fact, over 50\% of the frames experience a 
2--3 dB improvement in PSNR and a 0.05--0.0075 improvement in SSIM with both versions of \sys.

\begin{table}
\setlength\tabcolsep{3pt}
    \centering
    \small
    \begin{tabular}{l c c c c c}
    \toprule
\textbf{\#Feature Channels (F)} & \textbf{8} & \textbf{16} & \textbf{32} & \textbf{64} & \textbf{128} \\
         \midrule 
         PSNR(dB) &  38.49 & 38.69 & 39.87 & 39.89 & 39.90 \\
         SSIM &  0.942 & 0.944 & 0.946 & 0.947 & 0.949\\
         Inference Time (ms) & 7 & 9 & 11 & 17 & 25 \\
         Num. of Parameters & 0.59M & 1.14M & 2.22M & 4.39M & 8.72M\\
         \bottomrule
    \end{tabular}
    \vspace{3pt}
    \caption{Impact of the number of output feature channels in \sys'sadaptive convolutional
    block on inference time and quality 
    metrics for a small video snippet on a NVIDIA V100 GPU.}
    \label{tab:model speed}
\end{table}

\NewPara{Impact of number of Output Feature Channels.}
Since \sys downsamples frames at the encoder and then streams a model to the receiving client to resolve the decoded frames,
it is important that \sys performs inference fast enough to run at the framerate of the video on an edge-device with
limited processing power. %
Viewers need a frame rate that is at least
30 fps for good quality. Consequently, 
the inference time on a single frame cannot afford to be longer than 33ms.
In fact, the Meridian~\cite{meridian} video has a frame rate of 60 fps, so running low-latency inference is even more critical.

To evaluate the practicality of \sys's lightweight model, we evaluate the end-to-end
inference time per frame
on an NVIDIA V100 GPU as we vary 
the number of the output feature channels in the adaptive convolution block ($F$) in \Tab{model speed}. %
While increasing $F$ 
improves the PSNR and SSIM values due to better reconstruction of the fine details, 
it comes at a cost. 
With $F=64$ and $F=128$, the inference times of 17 ms and 25 ms respectively causing the
frame rate to drop below the input 60 fps. Further, the number of parameters increases to nearly $10M$, a steep number for the model to stream periodically.
As a result, we design \sys's model to use 32 output feature channels, ensuring it takes only 11 ms to run inference on a single frame.
In comparison, the EDSR generic SR model is about 5$\times$ slower to perform
inference on a single frame. Even the end-to-end neural 
video compression approach DVC~\cite{dvc} takes over hundreds of milliseconds to infer a single frame
at 1080p. %

\NewPara{Trade-off between \mb and \cb in \sys.}
The presence of a dedicated \ms and a \cs in \sys implies that the 
bitrate for each stream can be controlled independent
of the other, to achieve different compression levels. \Fig{model frame ablation} shows the impact of altering the \cb for a fixed \mb and vice-versa,
when encoding the Meridian video using \sys.
The content \bpp is varied by changing the quality (CRF) of the 480p H.265 stream.
In contrast, the contribution from the model \bpp is controlled by the fraction of
model parameters transmitted during each update. 

As anticipated, for a fixed amount of model \bpp corresponding to sending 1\% of the model parameters, 
PSNR improves as the \cb is increased. 
This is because as the quality of the underlying low-resolution H.265 frames improves, it becomes easier for the model
to resolve them
to their 1080p counterparts. Increasing the \cb from the lowest quality level of CRF 35 (with 0.0014 \bpp) to 
CRF 20 (with 0.003 \bpp) improves PSNR from 31 dB to 36 dB. However, increasing the \bpp for the content
beyond that yields diminishing returns on PSNR (also illustrated in \Fig{quality bpp vimeo}). 
At higher quality levels, \Fig{model frame ablation} suggest that modest increases in the \bpp allocated to the model 
result in large improvements to the PSNR. For instance, adapting 10\% of the model parameters consumes 0.006
\bpp, 6x more \bpp than adapting 0.5\% of the model parameters, but results in a PSNR improvement of 1dB from 36.31 dB to
37.32 dB.

\begin{table}
\setlength\tabcolsep{3pt}
    \centering
    \small
    \begin{tabular}{l c c c c c c}
    \toprule
         \textbf{Update Interval (s)} & \textbf{5} & \textbf{10} & \textbf{15} & \textbf{20} & $\mathbf{\infty}$ \\
         \midrule
         PSNR(dB)  & 37.25 & 36.52 & 36.57 & 36.45 & 35.32\\
         SSIM  & 0.92 & 0.91 & 0.91 & 0.91 & 0.91\\
         Bits-per-pixel & 0.006 & 0.003 & 0.002 & 0.0015 & 0 \\
         \bottomrule
    \end{tabular}
    \vspace{3pt}
    \caption{Impact of varying the interval over which \sys updates its model on the \bpp consumed by model updates and
    the associated gains in video quality. We find that an update interval of 5 seconds strikes a good trade-off
    between \bpp and quality.}
    \label{tab:update freq impact}
\end{table}

\NewPara{Impact of \sys's update interval.}
\adaptivesys can also control the \bpp consumed by the \ms
by varying the interval over which updates to the SR model are performed.
Frequent updates imply that
the \mb is higher, but the reconstruction is better since the model is trained on frames very similar to the current
frame. %
An extreme scenario is an update interval of $\infty$ that corresponds to the \ossys.
\Tab{update freq impact} captures the impact of varying the update interval on the average PSNR and SSIM of decoded frames from the Meridian video.
We find that an update interval of 5 seconds achieves good performance without compromising too much on \bpp. The fact
that the PSNR does not degrade significantly for modest increases to the update interval suggests further optimizations atop \sys
that only send model updates after a drastic scene change.

\section{Conclusion}
In this work, we present \sys, an approach that augments existing video codecs 
with a lightweight and content-adaptive super-resolution model. \sys
achieves video quality comparable to modern codecs with better
compression. Our design 
is a first step towards leveraging super-resolution as a video compression technique. Future work includes further optimizations to identify the pareto frontier for the model vs. \cb trade-off, more sophisticated techniques
to detect scene changes and optimize update intervals, as well as the design
of newer and more effective lightweight super-resolution neural network architectures.

{\small
\bibliographystyle{ieee_fullname}

\bibliography{egbib}
}

\end{document}